# A toolbox for calculating objective image properties in aesthetics research

Christoph Redies[1,*], Ralf Bartho[1,*], Lisa Koßmann[2], Branka Spehar[3], Ronald Hübner[4],

Johan Wagemans[2], and Gregor U. Hayn-Leichsenring[1]

[1] Experimental Aesthetics Group, Institute for Anatomy I, University Hospital Jena, Germany,
[2] Laboratory of Experimental Psychology, Department of Brain and Cognition, University of Leuven (KU Leuven), Belgium, [3] School of Psychology, University of New South Wales, Australia, and [4] Department of Psychology, University of Konstanz, Germany

* Equal contribution

Correspondence to:

Christoph Redies

Institute of Anatomy I

Teichgraben 7

D-07743 Jena, Germany

Phone: +49 – 9396 120

E-Mail: christoph.redies@med.uni-jena.de

**Abstract**

Over the past two decades, researchers in the field of visual aesthetics have studied numerous quantitative (objective) image properties and how they relate to visual aesthetic appreciation. However, results are difficult to compare between research groups. One reason is that researchers use different sets of image properties in their studies. But even if the same properties are used, the image pre-processing techniques may differ and often researchers use their own customized scripts to calculate the image properties. To provide greater accessibility and comparability of research results in visual experimental aesthetics, we developed an open-access and easy-to-use toolbox (called the "Aesthetics Toolbox"). The Toolbox allows users to calculate a well-defined set of quantitative image properties popular in contemporary research. The properties include lightness and color statistics, Fourier spectral properties, fractality, self-similarity, symmetry, as well as different entropy measures and CNN-based variances. Compatible with most devices, the Toolbox provides an intuitive click-and-drop web interface. In the Toolbox, we integrated the original scripts of four different research groups and translated them into Python 3. To ensure that results were consistent across analyses, we took care that results from the Python versions of the scripts were the same as those from the original scripts. The toolbox, detailed documentation, and a link to the cloud version are available via Github: https://github.com/RBartho/Aesthetics-Toolbox. In summary, we developed a toolbox that helps to standardize and simplify the calculation of quantitative image properties for visual aesthetics research.

Many contemporary models of aesthetic experience hypothesize that aesthetic judgments by human beholders are partially based on the perception of low- and mid-level visual features (Datta et al., 2006; Farzanfar & Walther, 2023; Gómez-Puerto et al., 2016; Graham & Redies, 2010; Ibarra et al., 2017; Li & Chen, 2009; Li & Zhang, 2020; Nakauchi et al., 2022; Palmer et al., 2013; Peng, 2022; Redies, 2015; Sidhu et al., 2018; Taylor et al., 2011; Zhang et al., 2021). Besides perceptual processing, cognition and emotions are considered to be crucial determinants of aesthetic experience ("aesthetic triad", Chatterjee & Vartanian, 2014; Hekkert, 2006). In support of the role of perceptual processes, quantitative (objective) image properties were found to predict aesthetic judgments such as ratings of liking, interest, and beauty on different types of visual stimuli, including artworks (for reviews, see Brachmann & Redies, 2017a; Chamberlain, 2022; Leder et al., 2004; Redies, 2007; Spehar et al., 2015; Taylor et al., 2011).

In the present article, we describe a toolbox (called "Aesthetics Toolbox") that can be used to compute a set of quantitative (objective) image properties previously studied in aesthetics research. These properties (e.g., self-similarity, entropy, or color features) are calculated based on the physical structure of digital images and are therefore independent of the beholder. By contrast, subjective image properties are based on the impressions or feelings that images evoke in beholders (Chamberlain, 2022; Lyssenko et al., 2016). Human responses to images can be measured, e.g., by psychological methods. When we refer to image properties in the present work, we mean quantitative (objective) image properties, unless stated otherwise. Note that, in the field of computer vision, image properties are usually referred to as image features.

We focus on properties that describe 2d static images, which have been more frequently studied than 3d or moving stimuli, such as architectural objects, sculpture, dance and movies (Christensen & Calvo-Merino, 2013; Joye, 2007; Vukadinovic & Markovic, 2012). Moreover, we will put emphasis on image properties that are derived from the perceptual structure of large image areas or from the entire image. Examples are Fourier spectral properties or the self-similarity of images (Aks & Sprott, 1996; Amirshahi et al., 2012; Graham & Field, 2007; Redies et al., 2007). Such image properties are of particular interest because many of the image features underlying subjective aesthetic impressions, such as "composition", "visual rightness" (Locher et al., 1999; Vissers & Wagemans, 2023), and "good Gestalt" (Arnheim, 1954), refer to global rather than to local image structure. For each image property, we will explain why it is relevant for aesthetics research and how it can be calculated. To contextualize the image properties, we will refer to some exemplary studies that illustrate their usage.

For the present work, we restrict our selection mostly to image properties that are known to us from our own research groups and for which we have the original code. Note that our coverage of image properties is not complete and there are other properties studied in experimental aesthetics. We do not attempt to give an exhaustive or general overview of this field of research in the present work. Rather, we aim to endow investigators with an understanding of how to calculate a subset of specific image properties and how to use them in aesthetics research. We do believe, however, that our choice of image properties is representative of the larger set and probably also includes the most frequently used ones.

Code to calculate many image properties for visual aesthetics research is freely available online, but it is scattered in various repositories. Examples are the toolbox by Mayer and Landwehr (2018), which computes four image properties related to processing fluency in the visual system (simplicity, symmetry, contrast, and self-similarity; Mayer & Landwehr, 2018). Walther et al. (2023) designed a series of computational tools to analyze mid-level visual properties, including local symmetry, contour properties and perceptual grouping cues in real-world images. Finally, to study the relation between order, complexity and aesthetic ratings, Van Geert and colleagues (2023) conceived a toolbox that allows researchers to create multi-element displays that vary qualitatively and quantitatively in different kinds and measures of order and complexity. Last but not least, Peng (2022) provided a tutorial on how to calculate image properties, such as color attributes and complexity, for research in the social sciences.

The advantage of the Aesthetics Toolbox is that it is a coherent, open access web application that is written in a single programming language, Python 3[1]. To our knowledge, it is the first toolbox that allows the calculation of image properties from multiple research groups on a single web interface. The Toolbox can be used without any proprietary software license. It is open source, fully documented, tested and maintained. All scripts in the toolbox are based on original code written by researchers in the field of aesthetics, ensuring that all image properties produce the same results as in the original studies for which they were developed. The code and the calculations methods are documented in detail. This allows researchers to replicate, compare, and extend the results of aesthetics research. The Toolbox has a browser-based GUI build with streamlit (https://streamlit.io) for easy usage and does not require programming knowledge. Finally, other research groups are invited to expand the Toolbox by contributing their own image properties and techniques to compute them. We thus hope that the Toolbox will become a valuable means to investigate objective image properties in aesthetics research.

## Objective image properties

This section provides a general introduction to each image property that can be calculated with the Toolbox. Additional computational details are documented online.[1] The content and text of the present description of the Toolbox partially overlap with the online documentation. We restricted our Toolbox to image properties, for which original code was available. For each image property, our Python code is based on the latest available version of the respective code. We confirmed that the code implemented in our Toolbox yields results identical to those from the original version of the code supplied to us, except for properties where this was not possible for technical reasons (HOG complexity and anisotropy). We will describe the image properties in the following sections, proceeding from relatively simple properties to more complex ones. Image properties are grouped according to which category of image property they reflect rather than by which method they are calculated. As a general convention, we capitalize the first word in the names of the image properties that are calculated with the Aesthetics Toolbox.

---

[1] https://github.com/RBartho/Aesthetics-Toolbox

**Image size, Scaling and Aspect ratio**

*Image size*

The calculation of image size and aspect ratio may seem relatively straightforward. To calculate Image size with the Toolbox, we follow Datta et al. (2006) and define Image size as the sum of image height and image width. With few exceptions, we do not use the product of height and width, i.e., the total number of pixels, as it follows a quadratic growth with increasing image height and width, quickly yielding very high numbers, which do not correlate well with the subjective impression of a steadily increasing image size.

Special attention should be given to the experimental conditions that affect image size, as well as to image properties that depend on image size. For example, the size of the images in the dataset of origin is not necessarily the same as the size used to collect ratings in an experiment where images may be scaled down for display. Moreover, the calculated values of several image properties depend on image size. Obvious examples are the measures that reflect image complexity. Other properties, such as the means of the color channels and their standard deviations, are less affected by image size. Therefore, it is advisable to use the same or a similar image resolution to calculate different image properties, especially when absolute values of the same image property are compared between images. We recommend a minimum length of 1024 pixels on the longer side of each image in order to avoid upscaling during pre-processing of images (for details on image pre-processing, see the documentation for individual image properties).

*Aspect ratio*

The methods for calculating aspect ratio lack consistency in the literature. Aspect ratio has been calculated as either the height-to-width ratio (e.g., see Mallon et al., 2014) or the width-to-height ratio (e.g., see Datta et al., 2006; Iigaya et al., 2021; Li et al., 2006). In the Toolbox, we follow the latter metric, which is the convention used for specifying display format in commercial settings. Both measures correlate negatively with each other.

Care should be taken to avoid changing the aspect ratio for the calculation of image properties. Changing the aspect ratio usually involves a change in image size. Moreover, changing the aspect ratio to obtain uniformly square images will necessarily affect the aesthetic appeal of images because the depicted objects or scenes may appear distorted.

**Contrast, Lightness entropy and Complexity**

*Root mean square (RMS) contrast*

Contrast is a commonly studied feature in aesthetics research. In the Toolbox, Contrast is calculated as the Root Mean Square (RMS) Contrast, which corresponds to the standard deviation of the Lightness (L) channel of the L*a*b* (or CIELAB) color space (Peli, 1990). Contrast is higher if pixel values are distributed over a larger range of Lightness values. There are several other methods to capture contrast (e.g., Li & Chen, 2009; Luo & Tang, 2008; Schifanella, 2015; Tong et al., 2004). It is unclear to what extent the different measures capture the same or a similar aspect of an image. Several studies have shown that images of high contrast are generally preferred over images of low contrast (e.g., see Tinio et al., 2011; van Dongen & Zijlmans, 2017).

*Lightness entropy*

Entropy is a concept used in various fields, ranging from thermodynamics to information theory. In general, low entropy is associated with highly organized, ordered structures, and the transition towards less organized states increases entropy. In information theory, Shannon entropy is a measure of the unpredictability of information content of a message source (Shannon, 1948). When applied to images, Shannon entropy measures the degree to which an image feature varies in an unpredictable fashion; it is inversely related to the notion of spatial redundancy (Kersten, 1987; Mather, 2018). A common way of expressing entropy of an image is with respect to the range of states or values that local samples of an image, such as pixels, can possess. These states are often summarized in histograms of the respective image property. The Toolbox includes measures for the entropy of (a) pixel intensities (this Section), (b) color hue (see Section "Colorfulness"), (c) the spatial distribution of pixel intensity (see Section "Homogeneity"), and (d) edge-orientation (see Section "Edge-orientation entropy").

The Aesthetics Toolbox calculates the Shannon entropy of pixel intensity based on the Lightness (L) channel of the L*a*b* (or CIELAB) color space. This measure has been simply referred to as 'entropy' by other researchers (Iigaya et al., 2021; Mather, 2018; Sidhu et al., 2018). To avoid confusion, we refer to it as Lightness entropy in the present work. Lightness entropy is high if all pixel values tend to occur with the same frequency in an image. It is low if particular pixel values occur more frequently than others.

*Complexity*

Image complexity can be broadly defined as the quantity and variety of information in an image. There are many different ways to calculate image complexity (for reviews, see Forsythe et al., 2011; McCormack & Gambardella, 2022; Nath et al., 2024; Peng, 2022; Van Geert & Wagemans, 2020, 2021). For example, methods for determining visual complexity have been based on the effectiveness of GIF or JPEG compression (Forsythe et al., 2011; Machado et al., 2015), edge density (Machado et al., 2015), fractal dimension (Bies, Blanc-Goldhammer, et al., 2016; Spehar et al., 2016; Taylor et al., 1999) and luminance and color gradients (Braun et al., 2013). On average, observers prefer visual stimuli of intermediate visual complexity (Berlyne, 1970; Taylor et al., 2005; Wundt, 1874), but there are large differences between individuals in which degree of complexity they find pleasing (Aitken, 1974; Bies, Blanc-Goldhammer, et al., 2016; Güclütürk et al., 2016; Spehar et al., 2016; Van Geert & Wagemans, 2021; Vissers et al., 2020).

In the Toolbox, we include two relatively straightforward measures of complexity that have been used in visual aesthetics research before. The first one is based on edge responses of Gabor-filtered images (here called Edge density; Mehrotra et al., 1992). The second measure is based on lightness and color gradients in an image (here called Complexity; Braun et al., 2013; Dalal & Triggs, 2005). Moreover, the Toolbox contains two additional measures, the Fractal dimension (Bies, Blanc-Goldhammer, et al., 2016; Spehar et al., 2016; Taylor et al., 1999) and the Fourier slope (Burton & Moorhead, 1987; Graham & Field, 2007; Redies et al., 2007; Spehar & Taylor, 2013; Tolhurst et al., 1992), which also relate to image complexity (Bies, Boydston, et al., 2016; Van Geert & Wagemans, 2020, 2021).

**Histogram of Oriented Gradients method.** To calculate Complexity values with the (Pyramid-)Histogram of Oriented Gradients ([P-]HOG) method (see Appendix in Braun et al., 2013; Figure 1; Dalal & Triggs, 2005), each colored image is converted into the L*a*b* color space and separated into its three channels. For each channel, a gradient image is calculated. The maximum gradient value of the three channels is then combined into a single gradient image (Figure 1 C). Because gradients are generally stronger for the L* (Luminance) channel of the L*a*b* color space, lightness gradients dominate the combined gradient image. The mean value over the combined gradient image is used as the measure of the Complexity of the image. A uniform original image with small changes in pixel values would result in a gradient image of low mean values, i.e., low Complexity, while an image with large changes would result in a gradient image of high mean values, i.e., high Complexity.

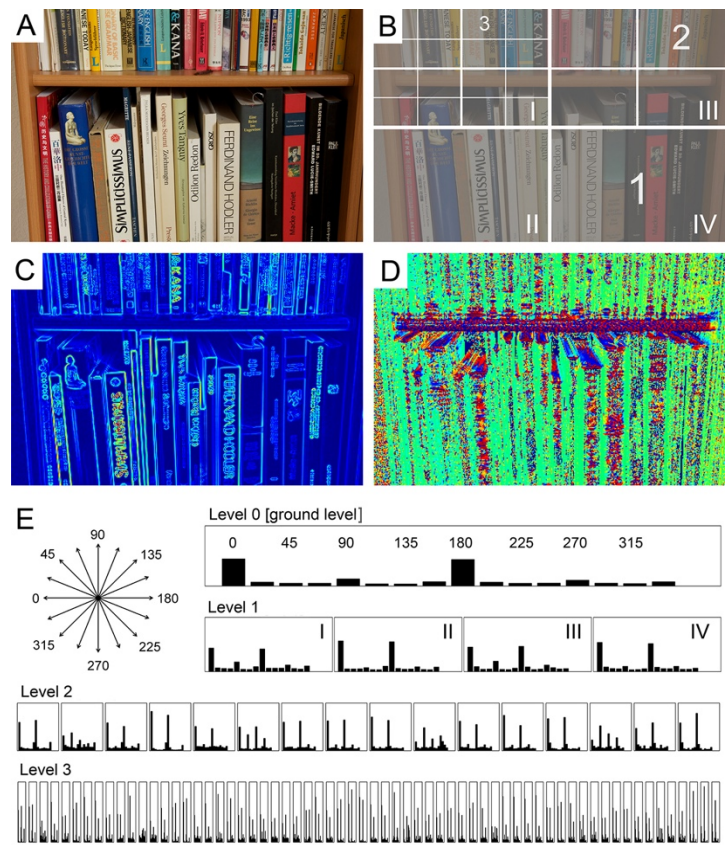

Figure 1. Calculation of Complexity, Anisotropy and Self-similarity by the (P-)HOG method, as explained in more detail in the text. (A) Original photograph. (B) Schematic diagram of the subdivisions at different levels of spatial resolution (Levels 1–3). (C) Pseudo-colored gradient strength image. (D) Image of pseudo-colored orientations (e.g., *red* for horizontal orientations, and *green* for vertical orientations). (E) The HOG features at the different levels of spatial resolution (Levels 0–3). The orientations of the 16 bins used for calculating the HOG features are displayed at the upper left side of the panel. Reproduced from Braun et al. (2017) with permission.

The Complexity values calculated by this method depend on image size (Redies & Gross, 2013). As mentioned above, it is thus advisable to use a uniform input size for all images analyzed. The desired image size can be set in the parameter section of this image property in the Toolbox.

**Edge density (Gabor filters).** As another measure of image complexity, we estimate the Edge density in Gabor-filtered images (Mehrotra et al., 1992), following the procedure described by Redies and colleagues (2017). In brief, the Toolbox starts by automatically converting the images to grayscale. Each input image is then reduced to a maximum size of 120,000 pixels (width $\times$ height). Next, we apply a bank of 24 oriented Gabor filters with equally spaced orientation bins to cover one full rotation (360°) in order to extract oriented edge elements from each image. Gabor filters resemble receptive fields at low levels of the human visual system. Edge density is calculated as the sum of all edge responses in each Gabor-filtered image. The resulting value reflects not only the density of edges in an image, but also their strength (i.e., contrast). Like other measures that are based on the number and strength of luminance gradients in an image (Forsythe et al., 2011), Edge density correlates with perceived (subjective) complexity.

**Color**

***Channel Means and Standard deviations***

Color is a pivotal feature of artworks and images in general (Bekhtereva & Muller, 2017; Li & Chen, 2009; Nakauchi et al., 2022; Nascimento et al., 2017; Palmer et al., 2013; Peng, 2022). To describe the color gamut of images, we capture color information in three widely used color models (RGB, HSV and L*a*b* [CIELAB]) that have been widely used in aesthetics research(Datta et al., 2006; Geller et al., 2022; Iigaya et al., 2021; Li & Chen, 2009; Li et al., 2006; Mallon et al., 2014; Nakauchi et al., 2022; Peng, 2022; Schifanella, 2015; Thieleking et al., 2020).

The RGB color space is one of the most popular ones; it is based on the primary colors of light and comprises a red (R) channel, a green (G) channel, and a blue (B) channel. There are different versions of this color space (e.g., sRBG and Adobe RGB). Differences in the color models used can have a large effect on the calculated image properties. Even if an image displayed on a screen does not differ subjectively for the human eye between the various color models, differences in the underlying data structures may strongly influence the results of the calculated image property. A perceptually more intuitive representation of the RGB color model is the HSV model (H, hue; S, saturation; and V, value).

The L*a*b* color model describes a perceptually uniform space that has a lightness or intensity (L*) channel and two color-opponent channels, a* (green-magenta channel) and b* (blue-yellow channel) where positive values are magenta and yellow, respectively. The Toolbox calculates Means and Standard deviations for each channel of the three-color models.

### *Colorfulness*

As a complement to the above color values, we calculate the Shannon entropy of the Hue channel of the HSV color space to capture the colorfulness of an image (Color Entropy; or HSV[H] entropy; see Geller et al., 2022). This measure shows high values if an image displays many color hues with about equal frequency across the entire range of hues, regardless of which colors these are in detail. An image with only a single hue would have a very low Shannon entropy in the hue channel.

### Symmetry and Balance

Balance is an attribute that reflects how well pictorial elements are arranged in the composition of an image and thereby contributes to its aesthetic appeal (Arnheim, 1954; Damiano et al., 2023; Hübner & Fillinger, 2016; McManus et al., 2011). Symmetry can be regarded as a special case of balance, to which the human brain is particularly sensitive (Bertamini et al., 2018; Jacobsen & Höfel, 2003). It is generally believed that symmetry is an important and universal basis of visual preference (for a review, see Bode et al., 2017;

Damiano et al., 2023). However, the universal role of symmetry as an 'aesthetic primitive' (Latto, 1995) has been contested (Leder et al., 2019).

To explain balance, Arnheim (1954) hypothesized that each rectangular frame possesses a field of invisible forces. Accordingly, the center of the framed image possesses the strongest attraction. The center is followed by its corners, the horizontal and vertical axis and then the diagonal axes. According to Arnheim, every element placed in an image is pulled by all the invisible forces stemming from the pixelwise structure and, additionally, by all other pictorial elements in the image, thus creating an inner tension (Hübner, 2022; Hübner & Fillinger, 2016; McManus et al., 2011). Related to these ideas, the Toolbox includes three different pixel-based ways to calculate geometric Balance and Symmetry, as proposed by Hübner et al. (2016). In addition, we describe a way to compute mirror symmetry based on features from low layers of a deep (convolutional) neural network (CNN) (Brachmann & Redies, 2016). These features are akin of filters in the early visual cortex and are thus more physiological than purely geometrical algorithms (see Section "CNN feature-based Symmetry").

### Pixel-based metrics

**Mirror Symmetry.** Two different approaches have been taken to measure geometric symmetry. One line of research focuses on the detection of symmetry and symmetry axes (e.g., Wagemans, 1995, 1997). Another line focuses on measuring the strength of symmetry in an image. Here, we will follow the latter line. While there are several forms of symmetry (reflectional, translational, and rotational symmetry; Liu et al., 2013), many researchers focus on one particular kind, namely reflectional symmetry ('mirror symmetry') which is a simple form of balance. Hübner and Fillinger (2016) defined this measure as the mean symmetry around the four main axes of an image (vertical, horizontal and orthogonal axes; see above). For quadratic images, a mean Mirror symmetry (MS) score is calculated from these four symmetries and expressed in percent. For other (non-quadratic) rectangular images, the mean

MS score is calculated for the vertical and horizontal axes only. Higher values indicate more symmetrical images.

**Balance.** Wilson and Chatterjee (2005) proposed a method to objectively measure balance in images of black geometric elements (circles, squares or hexagons) on a white square. They calculated eight symmetry measures by determining the number of black pixels per symmetric area of equal size around horizontal, vertical and diagonal axes, as well as on columns and lines. Mean symmetry values for all axes (here called "Balance score") that are closer to 0% reflect high Balance and those closer to 100% complete asymmetry. The authors showed that this objective measure of Balance correlates highly with subjective preferences (Wilson & Chatterjee, 2005). The Toolbox includes a version of the Balance score that works also on grayscale images. Here, the Symmetry values are the differences between the sum of all grayscale intensity values in the areas compared.

**Deviation of the Center of Mass (DCM).** A more complex characterization of balance is based on the concept of center of mass (McManus et al., 2011). Briefly, if the center of "perceptual mass" of all elements (depending on the perceptual weight and positioning of the respective element) is similar to the geometric center of the frame (i.e., the intersection between the horizontal and vertical axis in rectangular images), then the image is balanced (Arnheim, 1954; Hübner & Fillinger, 2016; McManus et al., 2011). Using this approach, Hübner and Fillinger (2016) studied black-and-white images of simple geometric patterns similar to those of Wilson and Chatterjee (2005). They assumed that the "perceptual mass" of a black pixel is 1, while that of a white pixel is 0. Based on this assumption, they computed the Euclidean distance of the center of perceptual mass to the geometric center of the image for each axis and expressed this distance as a percentage of the maximum possible distance from the geometric center of the given image (here called "DCM score"). Both of the above balance measures are expressed as percentages and are therefore easily comparable. In a rating study, DCM scores correlate more strongly with subjective balance ratings than the

Balance score, while APB scores correlated more strongly than DCM scores with subjective preference ratings (Hübner & Fillinger, 2016). The Toolbox implements a version of the score that also works on grayscale images (Thömmes & Hübner, 2018).

### *CNN feature-based Symmetry*

The three symmetry measures presented above are based on pixel values. Brachmann and

Redies (2016) developed a measure of symmetry that is based on filter responses from the

first layer of a convolutional neural network (CNN) (AlexNet; Krizhevsky et al., 2012). These

CNN features match responses of neurons in the early visual cortex of higher mammals (for reviews, see Bowers et al., 2022; Kriegeskorte, 2015; Lecun & Bengio, 1995; Lindsay, 2020;

Rafegas & Vanrell, 2016) and capture features such as oriented edges, color-opponent blobs,

and spatial frequency information (Figure 2). These features are reminiscent of the independent components of natural scenes (Bell & Sejnowski, 1997; Hyvärinen & Hoyer,

2001), and also emerge during the learning of a sparse code for natural images (Olshausen &

Field, 1996; for a review, see Simoncelli & Olshausen, 2001).

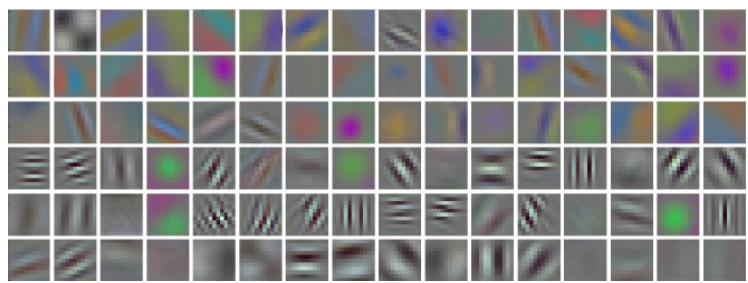

Figure 2. The 96 features on the first convolutional (conv1) layer of a CNN (AlexNet;

Krizhevsky et al., 2012). The filters respond to luminance edges of different orientations and spatial frequencies as well as color-opponent gradients. Reproduced from Brachmann et al.

(2017) with permission.

To measure higher-level symmetry with the Toolbox, we used first-layer (conv1) filter responses and computed Left-right symmetry, Up-down symmetry and a Symmetry combined for all four directions (left-right-up-down) according to the algorithm by Brachmann and Redies (2016). The authors showed that the CNN-based symmetry scores predict human perception of symmetry with high accuracy.

**Fractality and Self-similarity**

Natural scenes contain a high degree of regularity in their statistical structure, despite their considerable surface-level heterogeneity (Burton & Moorhead, 1987; Field, 1987; Ruderman, 1994; Tolhurst et al., 1992). It has been shown that large subsets of artworks resemble natural scenes in this respect (Graham & Field, 2007; Graham & Redies, 2010; Mather, 2018; Redies et al., 2007; Taylor et al., 2005). Because of this coincidence, natural scene statistics have been investigated in artworks and other types of aesthetically preferred images.

In both natural environments and artistic compositions, it is commonly observed that neighboring regions exhibit greater similarity in their spatial characteristics—for example textures, colors, orientation —compared to more distant regions. These consistent patterns are closely connected to concepts like scale invariance, self-similarity, and fractal scaling. Scale invariance is a feature of systems, statistics or objects that do not change properties if their scale changes by a certain amount, i.e., if one zooms in and out of an image. Self-similarity is a property in which a form is made up of parts similar to the whole or to one another. Fractal-like scaling properties are typically quantified using two distinct scaling methods: the slope ($\alpha$) of the Fourier amplitude spectrum ($1/f^{\alpha}$) and the fractal dimension ($D$). These two measures have been related both mathematically (Graham & Field, 2008) and empirically (Bies, Boydston, et al., 2016). Different methods to calculate fractal properties are implemented in the Toolbox. In addition, we describe two methods to calculate self-similarity by using features that more closely resemble neural responses filters in the early mammalian visual system (see Section "Self-similarity").

*Fourier spectrum Slope*

A common way to represent the spatial distance-dependent regularities regarding the intensity variations across natural scenes and other types of images is through the shape of their spatial frequency amplitude (Fourier) spectra. Figure 3A depicts the original versions and the spatial frequency-filtered versions of three different natural scenes. When the initial scenes are broken down into distinct spatial frequency components—depicted in the middle row (low spatial frequency-filtered images) and bottom row (high spatial frequency-filtered images)—it becomes evident that the relative amplitude of intensity variations is inversely correlated with the spatial frequency (f) content of these images. This relationship is illustrated in the right panel of Figure 3B. When plotted in the log-log coordinates, amplitude decreases linearly for most natural objects and scenes. The power-law relationship between the amplitude and spatial frequency, represented by the function $1/f^\alpha$, is characterized by the amplitude spectrum slope $\alpha$ which, on average, ranges from 0.8 to 1.5 (peaking at 1.2). For images of natural scenes, on average, this slope ranges from 0.8 to 1.5 (peaking at 1.2). If power is plotted instead of amplitude, the average peak of the power spectrum slope is about 2.4 (i.e., equal to twice the amplitude spectrum slope). The slope value is believed to indicate the scale invariance of natural scenes, suggesting that similar spatial structure can be observed as we zoom in or out across the coarse or fine spatial scales (i.e., low and high spatial frequencies, respectively). Several types of aesthetically preferred images, including traditional artworks, have slope values similar to those of natural scenes (Graham & Field, 2007; Graham & Redies, 2010; Koch et al., 2010; Mather, 2018; Redies et al., 2007).

Fourier slope can also be interpreted as a measure of complexity (Van Geert & Wagemans, 2020; Van Geert & Wagemans, 2021) because more shallow slopes (i.e., less negative slope values) are indicative of more fine detail in an image. Moreover, the Fourier spectra of images contain cues that can be used to objectively discriminate between angular and curvilinear stimuli (Watier, 2018). Curvilinear stimuli, which are generally preferred over angular stimuli (Bar & Neta, 2006; Gómez-Puerto et al., 2016), possess a lower number of peaks in plots that sum up the magnitudes over frequencies along specific orientations of the Fourier spectrum (Watier, 2024). In a similar vein, visually preferred images, such as artworks and graphic novels, are more homogeneous across the orientations of the Fourier spectrum than other types of stimuli, such as photographs of objects or natural scenes (Koch et al., 2010). These findings suggest that an isotropic magnitude profile of the Fourier spectrum across orientations may be indicative of visual preference in certain types of images.

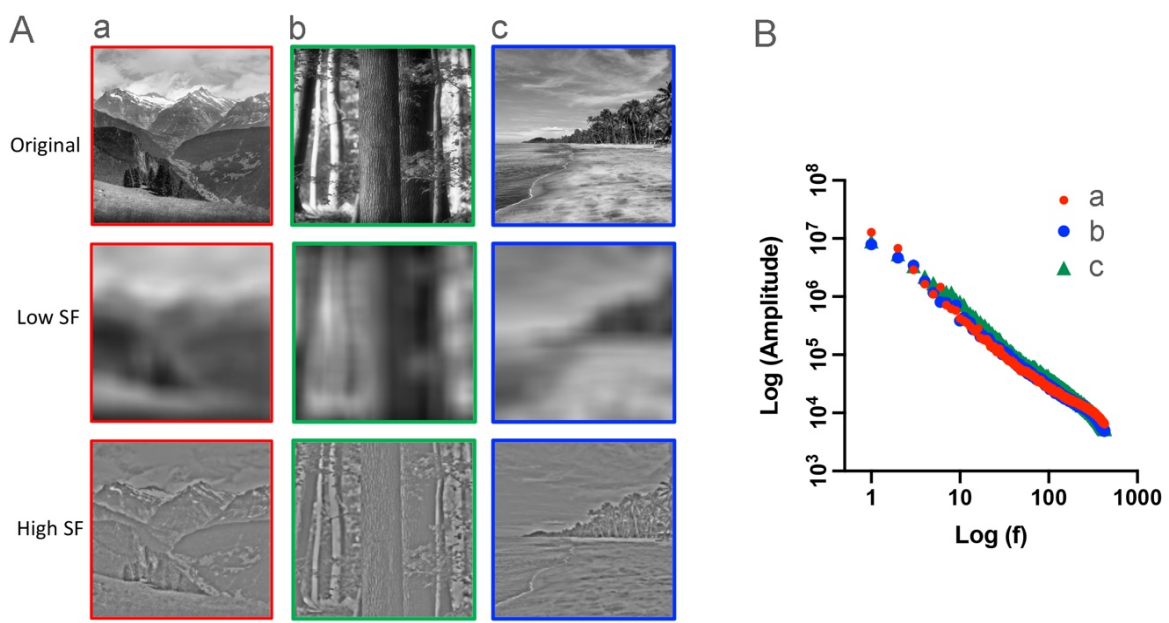

Figure 3A. Original images of natural scenes (top row) and each image filtered for low spatial frequencies (middle row) and high spatial frequencies (bottom row), respectively. B. Amplitude spectra of the original images. Abbreviation: SF, spatial frequency. Reproduced from Viengkham et al. (2019) with permission.

The Fourier spectrum slopes a have been calculated by many research groups (e.g., see Burton & Moorhead, 1987; Field, 1987; Graham & Field, 2007; Mather, 2018; O'Hare & Hibbard, 2011; Redies et al., 2007; Ruderman, 1994; Spehar & Taylor, 2013; Spehar et al., 2016; Tolhurst et al., 1992). There are several implementations of the algorithm that are similar but differ in important details. The general approach is to first Fourier transform the grayscale version of the image. In a log-log plot of the radially averaged Fourier spectrum (Figure 3B), the decrease in spectral power (or amplitude) is then approximated by a linear regression. The (negative) slope of this regression line is the Fourier spectrum Slope $\alpha$.

Specific implementations of these Fourier measures can vary in details of image pre-processing, the choice of the spectrum plot (amplitude or power), the basis of the logarithm used, and which frequencies should be excluded from the fitting because they represent noise or are prone to artefacts, such as uneven illumination, rectangular sampling, raster screen or noise in the images. Seemingly minor differences in implementation details can cause substantial deviations in absolute terms and in the between-group correlations of the Slope values.

In the Toolbox, we include three different versions for calculating the Fourier Slope. They are the methods by Branka Spehar and colleagues (Spehar & Taylor, 2013), by Christoph Redies' group (Koch et al., 2010; Redies et al., 2007), and by George Mather (Mather, 2014). Major differences between these versions are listed in Table 1.

Table 1. Characteristics of three methods to calculate the Slope values of the Fourier (amplitude or power) spectrum.

| | Code provided by | | |
| --- | --- | --- | --- |
| | Branka Spehar's Lab | Christoph Redies' Lab | George Mather |
| Pixel values | 8-bit grayscale | 8-bit grayscale | L (Lightness) channel of the MATLAB CIELAB color space |
| Image pre-processing and image format | Center crop to largest square with power of 2 (unless square already) | Padding images to square with mean gray value and resizing to $1024 \times 1024$ pixels | Center crop to largest square with power of 2 and resizing to $1024 \times 1024$ pixels |
| Fourier spectrum | Amplitude | Power | Amplitude |
| Frequency omitted from fitting | Frequencies with Cook's distance > n/4 | Frequencies below 10 and above 256 cycles/image | Lowest quartile and highest quartile of frequencies |
| Binning of frequencies | None | Binning of data points at regular intervals in log-log plot | None |
| Reference(s) | Isherwood et al. (2021) | Redies et al. (2007); Koch et al. (2010) | Mather (2014) |

To assess how similar the Slopes values are between methods, we determined the Pearson's coefficients of correlation for three sets of images: (a) 1000 random selected images from the AVA dataset (Murray et al., 2012), (b) the JenAesthetics dataset of 1629 traditional Western paintings (Amirshahi et al., 2015), and (c) 1000 artificially generated random-phase images (Branka Spehar, unpublished data). Pearson's coefficients for correlation of the Slope values calculated with the different methods were 0.76-0.86 for the AVA dataset, 0.67-0.88 for the JenAesthetics dataset, and 0.98-1.00 for the random-phase dataset. Slope values obtained with the three methods are therefore not comparable between methods for all data sets. However, for the artificially generated random-phase images, the results correlate almost perfectly. Because it is hardly possible to decide which method is better or worse from a theoretical point of view, we included all three versions in the Toolbox so that users can take their choice.

Ambiguities in calculating the image properties, as exemplified by the Fourier Slope, underline the need to exchange and share algorithms if the results from different studies are to be compared. Ideally, a standardized set of algorithms should be used by researchers in the field. The reimplementation of algorithms based on sketchy descriptions in papers can only be discouraged.

### Fourier spectrum Sigma

Another measure that can be derived from the log-log plot of Fourier power *versus* spatial frequency is the deviation of the data points from the regression line. This measure, which has been called Fourier Sigma (Redies et al., 2020), is calculated as the sum of the squares of the deviations, divided by the number of data points. Typically, the data points are sampled at equally spaced intervals in the log-log plots to avoid an overrepresentation of high frequencies (Table 1). The Sigma value indicates how linearly the log-log plot of the Fourier spectrum decreases with increasing spatial frequency. Higher values of Fourier Sigma correspond to larger deviations from a linear course. Sigma values are generally low for natural images and large sets of traditional artworks, i.e., a straight line fits the log-log plots well (Graham & Field, 2007; Koch et al., 2010; Mather, 2014; Redies et al., 2007). Images perceived as unpleasant typically show deviations of the spectral curve from a straight line (Fernandez & Wilkins, 2008; O'Hare & Hibbard, 2011).

*Fractal dimension (2d and 3d)*

The concept of scale invariance in natural scenes and patterns can also be quantified using a geometric scaling parameter called the **fractal dimension (*D*)**. Specifically, the fractal dimension examines the boundary edges between filled and empty regions within an image. Naturally occurring fractal structures, such as branches, ferns or diverse growth patterns, have been studied since the 1970s. Fractals exhibit recurring patterns that become increasingly complex at finer scales, culminating in scale-invariant shapes of remarkable complexity (Mandelbrot, 1983). The discovery that the abstract drip paintings by Jackson Pollock have a fractal structure (Taylor, 2002) was one of the first examples of an objective and complex image property measured in artworks.

The box-counting technique is commonly used to determine the fractal dimension (as shown in Figure 4). This method involves overlaying an image with a grid of equally-sized squares (referred to as 'boxes') of varying side lengths (*L*) and counting the number of squares, denoted as *N*, that intersect with the boundary edge. This count is repeated for increasingly smaller squares within the grid. Reducing the box size (i.e., using smaller values of *L*) is equivalent to examining the image at finer spatial frequencies. The quantity *N* represents the amount of space occupied by pattern boundaries at different spatial scales. The scale-invariance of the fractal pattern becomes evident through a power-law relationship. In this context, the exponent *D* represents the fractal dimension, which can be quantified by plotting *log N* as a function of *log (1/L)*.

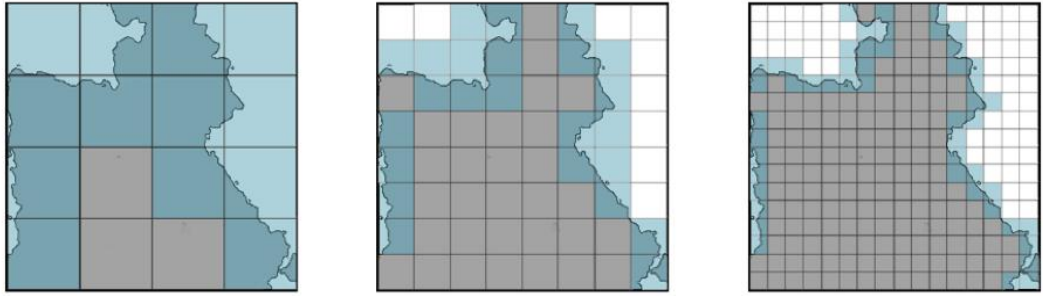

Figure 4. Illustration of the box counting technique for three different box side lengths (*L*). As L decreases from left to right, the number of boxes (*N*) required to measure the length of a boundary edge increases, following a power-law relationship defined by the fractal dimension *D*. Reproduced from Viengkham et al. (2019) with permission.

Fractals characterized by a low *D* value exhibit a repetition of patterns across varying scales, yielding a remarkably sleek and minimalist form. Conversely, fractals with a high *D* value showcase intricate and elaborate structures emerging from the repetition of patterns, imbuing the shape with complexity and detail (Taylor & Sprott, 2008). *D* is an index of the relationship between the coarse and fine geometrical structure in a repeating pattern and, as such, it is an image property that can be used to quantify the complexity of a given fractal pattern in an image. It is important to emphasize that it is thus not a numeric tool to detect fractal patterns in the first place. In support of this notion, it has been found that the perceived complexity of a wide range of images (ranging from natural to paintings and synthetic patterns) all increase linearly with increasing *D* values (Bies, Blanc-Goldhammer, et al., 2016; Forsythe et al., 2011; Mureika & Taylor, 2013; Spehar et al., 2003; Viengkham & Spehar, 2018). Interestingly, *D* is inversely related to the Fourier spectrum Slope value, another measure that reflects the complexity of an image (see Section "Fourier spectrum Slope"). A higher (negative) Slope value, i.e., a steeper slope, is equivalent to a lower *D* value, and vice versa. A detailed comparison of the two techniques can be found elsewhere (Bies, Blanc-Goldhammer, et al., 2016; Fairbanks & Taylor, 2010; Spehar & Taylor, 2013).

The Toolbox contains two different methods to calculate $D$, the 2d algorithm used by Branka Spehar and colleagues (Viengkham & Spehar, 2018) and the 3d algorithm used by George Mather (Mather, 2018). The main difference is that, for the 2d $D$, the image is binarized; the additional dimension for the 3d $D$ is the Lightness (L) channel of the CIELAB color space. Note that both versions correlate well, with a Pearson's coefficient $r = 1.00$ for 1000 synthetic random-phase images (Branka Spehar, unpublished data), 0.62 for the JenAesthetic dataset of Western paintings (Amirshahi et al., 2015), and 0.68 for 1000 random images from the AVA dataset of artworks (Murray et al., 2012).

To calculate the 2d $D$ (Viengkham & Spehar, 2018), the algorithm converts images into binary or edge-only versions of images (Figure 5). It is important to emphasize that the calculation of 2d $D$ is independent of photometric properties of images, such as perceived brightness. Instead, it is based on the degree of spatial variations along the edges of binarized, black-and-white regions in an image. This is the reason why before a box-counting procedure is applied, images are first thresholded with respect to their mean lightness (or pixel intensity), resulting in a two-tone image as illustrated in Figure 5B. Edge-only image variations of algorithms to calculate $D$ (Redies et al., 2015) are based on edge extraction from the thresholded black-and-white images (Figure 5C), e.g., by using the Canny algorithm (Canny, 1986). Possible values for 2d $D$ are between 1 (low complexity) and 2 (high complexity).

The binary image is analyzed at different scaling levels using the box count method, as described above, where the smallest size of each window is $2 \times 2$ pixels. A log-log plot with base 2 is used to fit a linear regression between the scaling level of the image (measured in side length) and the number of edges in the respective images. The slope of this regression line corresponds to the 2d *D*.

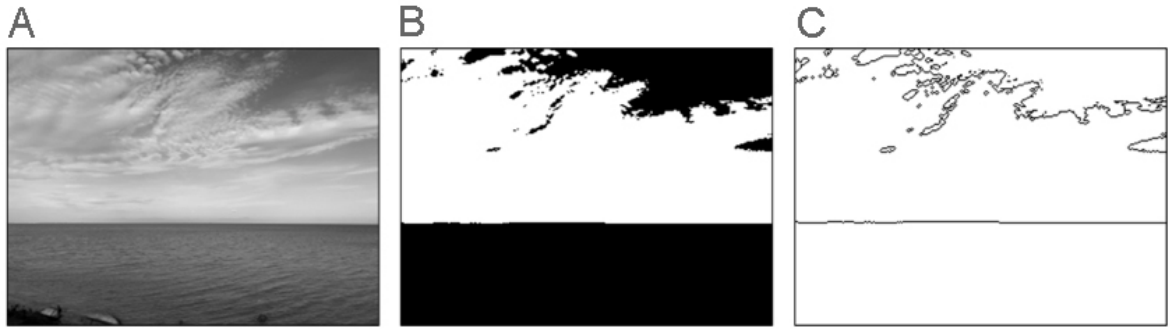

Figure 5: Original grayscale image (A), thresholded image (B), and edges-only image (C).

However, despite undergoing thresholding and edge extraction, the inherent fractal-scaling characteristics of images are preserved. This consistency renders these types of images and transformation techniques particularly appropriate for the study of the perception and aesthetics of fractal patterns across various image categories (Spehar & Taylor, 2013; Spehar et al., 2016; Viengkham & Spehar, 2018).

To measure the 3d *D* (Mather, 2018), image pre-processing consists of cropping each image to the largest central square region, which is then transformed into the CIELAB color space. From the Lightness channel of this color space, the fractal dimension is calculated using a 3d box-counting algorithm, which keeps the size of the image constant while changing the size of the boxes. This is unlike the 2d algorithm that keeps the size of the boxes constant as the image is resized to different magnifications (see above). The count of each individual box is the difference between the highest and lowest pixel lightness value. The 3d version of calculating *D* is based on a log-log plot of box size versus the sum of the counts of each box.

The logarithm used here has base E. As for the 2d version, *D* corresponds to the slope of the regression line in this plot. The 3d *D* value has a range from 2 (low complexity) to 3 (high complexity).

### *Self-similarity*

In general, self-similarity is a measure of how similar smaller subregions of an image are to the entire image with respect to specific image properties. Closely related concepts are scale invariance and fractality. The establishment of methods for measuring self-similarity was motivated by previous studies, which showed that large subsets of aesthetic images, including artworks, have a self-similar structure (Taylor et al., 1999, 2011). In particular, many visually pleasing images exhibit scale-invariance in the Fourier domain, which they share with images of natural scenes (Alvarez-Ramirez et al., 2008; Graham & Field, 2007; Redies et al., 2007). These results suggest that the spectral properties of large subsets of aesthetic images are self-similar at different levels of spatial resolution (scale invariance; Graham & Redies, 2010; Koch et al., 2010).

Alternative measures of self-similarity have been developed that can be related more directly to models of visual system function. In the Toolbox, we include two such measures. One is based on histograms of oriented gradients (HOG features, Dalal & Triggs, 2005) and the other uses low-level CNN features.

**(P)HOG-based Self-similarity.** In this method, Self-similarity is derived from the Pyramid Histogram of Oriented Gradients (PHOG) descriptor (Amirshahi et al., 2012; Bosch et al., 2007). The method was originally developed for image classification and to describe spatial shape (Dalal & Triggs, 2005). A step-by-step explanation of the method can be found in the appendix to the publication by Braun et al. (2013).

Because the values calculated for Self-similarity depend on image size (see Figure 7 in Redies and Groß, 2013), size should be normalized before PHOG analysis, e.g., to a total of 100,000 pixels (width × height, Braun et al., 2013). Resizing can be carried out with the separate function of the Toolbox or in the popup menu of the PHOG method.

To calculate Self-similarity, images are first converted into the L*a*b* color space and a gradient image is calculated for each channel, as described in Section "Complexity". Then, a PHOG descriptor is generated for each image based on HOG features that were obtained for equally sized bins and cover the full circle (360 degrees; Figure 1). Commonly, 8 or 16 bins that cover 180 or 360 degrees, respectively, are used. With 180 degrees (half a circle), the direction of the oriented gradients is ignored in the calculation. Initially, the HOG features are calculated for the entire image at the ground level (Level 0). Then, HOG features are calculated for consecutive levels of an image pyramid (i.e., PHOG, \Bosch et al., 2007) up to a given level (Figure 1). A pyramid is obtained by dividing the ground image into four rectangles of equal size (2 × 2 grid; Level 1). Each section at Level 1 is then divided again into four rectangles of equal size to obtain the next level of the pyramid, and so on. Accordingly, Level 2 contains 16 sections (4 × 4 grid) and Level 3 contains 64 sections (8 × 8 grid). Typically, HOG features are generated up to Level 3 in previous studies, which is also the top level for the analysis in the Toolbox.

The HOG features at Levels 1, 2 and 3, respectively, are compared to the HOG feature at the ground level (Figure 1) with the Histogram Intersection Kernel. In the popup menu of the Toolbox, it can be specified which levels are included in the calculation and with what weight. For example, Self-similarity has been expressed as the mean value for Levels 1-3 of the pyramid, with equal weight for all levels (Redies & Brachmann, 2017; Redies et al., 2020). At higher levels, the calculated values become unstable, because the image sections used in the analysis are exceedingly small and the HOG features become noisy (Amirshahi et al., 2012). Self-similarity is higher (closer to 1) if the HOG features at different levels of the pyramid are more similar to the ground level HOG feature. Low values that approach 0 indicate minimal Self-similarity between the HOG features. Traditional paintings and other visually pleasing stimuli have an intermediate to high degree of Self-similarity (Braun et al., 2013).

**CNN-based Self-similarity**. In Section "CNN feature-based Symmetry", we introduced Convolutional Neural Networks (CNNs) as models of human visual function at early convolutional layers (Figure 2). We argued above that, for symmetry detection, low-level CNN features are better suited to model symmetry detection by humans than pixel-based methods that focus on specific image properties, such as luminance gradients in the PHOG method (see Section "Histogram of Oriented Gradients [HOG] method"). A similar case can be made for Self-similarity. Brachmann and Redies (2017) developed a CNN-based method to measure Self-similarity by simultaneously considering information about luminance edges, color, and spatial frequencies. The authors showed that the CNN-based method yielded results, which resemble human visual function more closely than those obtained with the PHOG-based method (Brachmann & Redies, 2017b). To measure Self-similarity by the CNN-based method, the network processes the image to generate the filter responses for the first convolutional layer of an AlexNet pretrained on an ImageNet (Krizhevsky et al., 2012). From these results, histograms of maximum responses are produced over a grid of equally sized subsections of an image. Similar to the PHOG-based method, this histogram is then compared to the histograms of 64 subsections ($8 \times 8$ grid). Self-similarity is calculated as the median of all calculated values. A value closer to 1 indicates higher Self-similarity and a value closer to 0 lower Self-similarity.

**Feature distribution and entropy**

As the last group of image properties, we will present additional algorithms that focus on the distribution of image features, such as oriented gradients or edges, across an image. We will conclude with additional image properties that are derived from CNN-based features.

*Homogeneity*

In Section "Lightness entropy", we introduced the entropy of lightness (pixel intensity) to determine how uniform the frequencies of Lightness values are in an image. This measure was based on frequency histograms across all possible lightness values in an image. Here, we consider Homogeneity, an image property similar to lightness entropy (Hübner & Fillinger, 2016). To compute this measure, the input image is first converted into a binary (black-and-white) image using the Otsu method (Otsu, 1979). The binary image is then divided into 10 ✕ 10 sub-regions of equal size. For each sub-region, the number of black pixels is determined. The resulting 10 ✕ 10 matrix contains the number of black pixels per sub-region. These values are now summed, once for the rows and once for the columns of the matrix, resulting in two histograms for the vertical sum and the horizontal sum, respectively. For both histograms, the Shannon entropy is calculated and divided by the maximum possible entropy. Results are scaled by the factor of 100, which converts the two values into percentages. The value for Homogeneity is the average of the horizontal and vertical values of relative entropy. As a percentage, Homogeneity can take values from 0 to 100. An image with exactly the same quantity of black pixels in each of the 10 ✕ 10 sub-regions would have maximum Homogeneity, while an image with black pixels in one sub-region only and none in the others would have a minimum Homogeneity value. Unlike Lightness entropy, Homogeneity also captures the spatial distribution of pixels. For an image that has all black pixels in a sub-region, Lightness entropy can still be relatively high, but Homogeneity is not.

*Anisotropy of gradient orientation*

Above, we described measures of Complexity and Self-similarity that can be derived from the Histograms of Oriented Gradients (HOG) descriptor (Dalal & Triggs, 2005). The same descriptor can also be used to calculate how different gradient strengths are distributed across orientations in an image, a property here called Anisotropy (Braun et al., 2013). To calculate this property, the representation of lightness and color gradients of an image is mapped in a so-called gradient image. From this image, histograms of the oriented gradients (HOGs) are then generated for equal-sized orientation bins and subregions of the image. In the Toolbox, each image is divided into 64 subregions at Level 3 of the HOG pyramid (see Section "Histogram of Oriented Gradients method"). Anisotropy is calculated as the standard deviation between the summed and normalized gradient strengths for each orientation bin of each subregion (Braun et al., 2013).

If the histogram entries become more uniform across orientations, that is, if all orientations tend to be of equal strength in each subregion, Anisotropy decreases with a lower limit of zero. If specific orientations predominate in subregions of an image, e.g., horizontal and vertical orientations in the photograph of a building facade, histogram entries are more heterogeneous, and Anisotropy is high. In general, traditional Western artworks tend to be less anisotropic than diverse types of non-art images (Melmer et al., 2013; Redies et al., 2012).

### *Edge-orientation entropy*

Next, we describe a method to study the distribution of edge orientations in an image. This method is based on edge-filtered images and refers to the probability of encountering a particular orientation in an image (Geisler et al., 2001). First-order edge-orientation entropy (1st-order EOE) is a measure of how uniformly the edge orientations are distributed across the full spectrum of orientations in each image; it is thus indifferent to the arrangement of the individual edges in the image. Second-order EOE is a measure of how independent the spatial positions of edge orientations are across an image; it captures the spatial arrangement of the edges in the image (Geisler et al., 2001; Redies et al., 2017).

The motivation for using these measures in aesthetics research was twofold. First, the perceptual analysis of the spatial layout of contours and edges plays an important role in contour grouping and object recognition in natural images (Geisler et al., 2001). For artworks, some artists and art theorists have argued that the spatial layout of pictorial elements ("composition") is an important determinant of visual appreciation (Arnheim, 1954; Locher et al., 1999; Redies et al., 2017). Indeed, large subsets of artworks of different cultural provenance display high EOE values (Redies et al., 2017). Second, the distribution of edges across orientation histograms carries information about how angular or curved stimuli are (Grebenkina et al., 2018; Stanischewski et al., 2020; Watier, 2024). In general, curved stimuli have a more homogenous edge-orientation histogram than angular ones and are aesthetically preferred (Bar & Neta, 2006; Bertamini et al., 2016; Chuquichambi et al., 2022). Like other measures of entropy, EOE is a measure of how predictable a given feature is (see also Section "Lightness entropy"; and, for color entropy, see Section "Colorfulness").

For both EOEs, the Toolbox extracts edges by applying a bank of 24 oriented Gabor filters, which covered one full rotation when combined. Gabor filters are akin to simple cell responses in the primary visual cortex (Mehrotra et al., 1992). By applying oriented Gabor filters to an image, summary statistics of edge orientations in an image can be obtained. Due to computational limitations, we limit the number of edges for pairwise analysis to the 10,000 strongest edge responses for the calculation of the 2nd-order EOE. Despite this limitation, 2nd-order EOE is by far the computationally most expensive image property in the Toolbox.

First-order EOE is defined as the Shannon entropy of the summary orientation histogram that represents the full spectrum of edge orientations for the entire image (Redies et al., 2017). Entropy is high for uniform orientation histograms, i.e., if all orientations are present at about equal strength in the image, such as for most curved stimuli. It is lower for unevenly distributed histograms, as is the case for angular stimuli in general. Large subsets of traditional paintings have relatively high 1st-order entropy (Redies et al., 2017), indicating that they have relatively homogenous orientation histograms, as also suggested by their Fourier spectra (Koch et al., 2010).

To obtain 2nd-order EOE, the orientation of each strong edge in an image is related pairwise to the orientation of other strong edges in an image (Geisler et al., 2001; Redies et al., 2017). To avoid local regularities, such as collinearity, edge pairs of less than 20 pixels distance are excluded from the analysis. Histograms of the orientation differences are then obtained for all (strong) edge pairs in an image. Second-order EOE is maximal if all orientation differences occur at equal strengths in the histograms. In that case, edge orientations are distributed independent of each other in the image, i.e., edge orientation at one position in an image does not allow to predict the orientation at other positions. Examples for such images are photographs of particular types of natural objects, such as lichen growth patterns, but also man-made patterns, such as synthetic line patterns, decorated building facades or artworks of different cultural provenance (Geller et al., 2022; Grebenkina et al., 2018; Redies et al., 2017; Stanischewski et al., 2020). Note that for 2nd-order EOE to be high, 1st-order EOE must be high as well (Redies et al., 2017).

### *CNN feature variances*

In previous sections, we derived two image properties from CNN features, Symmetry, and Self-similarity. As mentioned above, features from the first layers of some CNNs, like the AlexNet used here (Krizhevsky et al., 2012), resemble neural responses in the early human visual cortex in that they process different types of information, such as oriented edges, color-opponent blobs, and spatial frequency information simultaneously (Kriegeskorte, 2015; Rafegas & Vanrell, 2017). So far, we have described image properties that reflect individual aspects of image structure independently from each other, without accounting for overall similarities and interactions in their response statistics, such as their variations in frequency and spatial distribution. For a more comprehensive analysis of image features, Brachmann and colleagues (2017) proposed to calculate two types of response variances of CNN filter responses, i.e., Sparseness and Variability, at the first convolutional layer of a CNN (conv1 of the AlexNet).

Layer conv1 comprises 96 response filter maps (Figure 2). About half of these maps display color-opponent characteristics, so that the CNN variances reflect color information more comprehensively than most of the other image properties (e.g., the Fourier slope or the [P-]HOG-based measures). For our calculations, each map is partitioned into $n \times n$ subregions of equal size. Responses are then recorded for every filter in each subregion by a max-pooling operation over the response maps. Two types of variances of the CNN features are calculated. First, we calculate the total variance for each feature over all 96 filter entries of the $n \times n$ subregions of conv1. This variance assumes high values if there is a low number of responses in a small number of subregions. It can thus be interpreted as the *Sparseness* of filter responses. Conversely, it is low if a large number of filters respond at similar strength at many image positions (*Richness*).

Second, we calculate the median over the variances of each of the 96 filters, again for different numbers of $n \times n$ subregions at conv1. This variance is high if there is a high degree of variability of filter responses across the response maps, whereas lower values suggest less variability of filter responses or higher Self-similarity (Brachmann et al., 2017).

To determine the two variances at different levels of spatial resolution, the Toolbox offers to select the number of $n \times n$ subregions at conv1, typically between $2 \times 2$ and $30 \times 30$ subregions. Because values for Sparseness and Variability, respectively, tend to correlate strongly between different levels of resolution, analyzing each variance at one representative resolution often suffices to obtain representative results.

The two variances diverge for different types of natural and man-made images (Brachmann et al., 2017). Large subsets of traditional artworks of diverse cultural provenance (Western, Islamic, and Chinese) are characterized by a high Richness (or low Sparseness) of filter responses and an intermediate to high degree of Variability (or Self-similarity; Figure 6). Other types of images, e.g., photographs of simple objects, urban scenes, natural scenes and plant patterns, exhibit different ranges for the two variances, and there is little overlap with the traditional artworks (Figure 6).

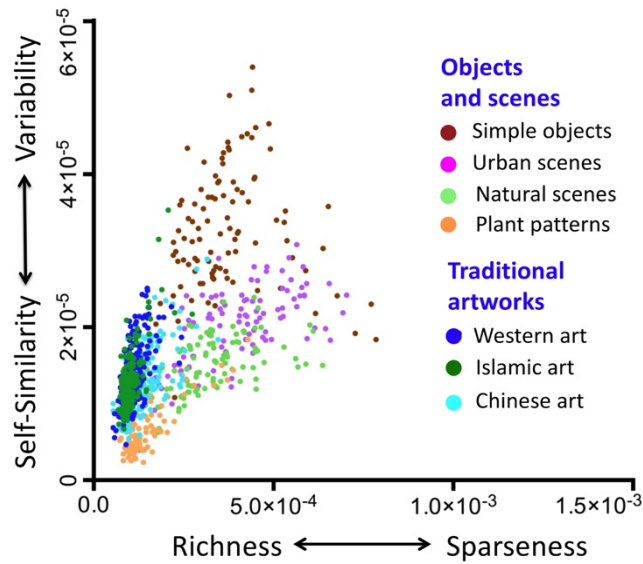

Figure 6. Dot plot of variances of CNN feature responses for traditional artworks from different cultural provenance and non-art stimuli, such as photographs of simple objects, urban scenes, natural scenes and plant patterns. Reproduced from Brachmann et al. (2017) with permission.

## Technical Considerations

### Translation of original scripts to Python 3

Original scripts to compute image properties with the Toolbox were gathered from various researchers, including Ronald Hübner (University of Konstanz), George Mather (University of Sussex), Branka Spehar (University of New South Wales), and the Experimental Aesthetic Group at the University of Jena. Many of these scripts were originally written in Python 2 or MATLAB. Python 2 has not been maintained since January 1, 2020; the current Python version is 3.12 as of March 2024. Python 2 is no longer supported by most modern Python packages. MATLAB is a proprietary software that requires a paid license. Furthermore, the built-in functions of MATLAB are not open source and the exact structure of its algorithms is thus not transparent to the user. Therefore, we translated all of the scripts to Python 3.

Care was taken to ensure that the Python 3 versions produced the same results as the original Python 2 and MATLAB scripts. A near perfect match was accomplished for all image properties, except for HOG Complexity and Anisotropy. For these two properties, the original MATLAB implementation allowed users to automatically scale the images prior to calculation. It was not possible to transfer this resizing operation one-to-one to Python 3 because the MATLAB *resize* function does not return the same results as similar Python implementations (e.g., Pillow, scikit-image, or opencv, see also Watier [2024]). Furthermore, MATLAB source code is also not easily reverse-engineered into Python because it is not open source. Without initial resizing, however, the Toolbox version of these image properties gives the same results as the original MATLAB version. Therefore, resizing is still possible in the Toolbox version, but the results are different from those of the original MATLAB versions.

In most of the original versions of the scripts, the documentation provided in the code was rather limited. In the new Python 3 versions, extensive documentation is available for each individual image property. Nevertheless, the structure and form of the novel scripts (names of functions, classes, etc.) still reflect the original implementations. The Github website contains all Python 3 scripts for calculating the image properties and the graphical user interfaces that were built with streamlit. It also comprises the original MATLAB and Python 2 scripts.

In the following sections, we will point out several technical issues that more generally pertain to calculating image properties and to running the Toolbox in an efficient way.

**Running the Toolbox efficiently**

*System requirements*

The Toolbox runs natively in the browser on all end devices. There are two ways in which the Toolbox can be used: First, the cloud version is available at https://aesthetics-toolbox.streamlit.app/ and allows the toolbox to be used without local installation, but with limited resources, i.e., the memory and computing power are comparatively low here. This is why the cloud version is primarily suitable for calculating image properties for small data sets and for testing the toolbox. With the cloud version, the images have to be uploaded to the Streamlit Community Cloud and calculated there, which can be problematic in terms of image copyrights. Second, the Toolbox can be installed locally. The maximum number of images that can be calculated simultaneously and the computing time depend on the resources of the local system. With the local installation, the browser is used only as an interface, but no data is uploaded to the internet.

*Multithreading*

Ideally, in view of the increasing amount of data to be analyzed, it would be desirable for the Toolbox to support multithreading on modern multi-core CPUs. Unfortunately, the common operating systems (Mac OS, Linux, and Windows) handle multithreading differently, so that the Toolbox's requirements for performance collide with its goal of platform independence, which was prioritized for the design of the Toolbox. For large data sets (larger than 6 GB), the user should install the local Toolbox version and divide the images to be calculated into several sets and start a separate local instance of the application in the browser for each set. In the background, the respective operating system will automatically assign the instances of the Toolbox that run in parallel to the logical CPU threads. The Github site also contains a script-only (no GUI) version of the Toolbox that can be used for local multi-threading or deployment to an HPC cluster.

*Computational load: Dependence on image size and image property*

Two important factors in the performance of the Toolbox are the size of the images and the image properties selected. The number of pixels in an image increases with the square of its side length. Overall, the computation time required for most of the image properties in the Toolbox increases exponentially with the resolution of the images.

A good heuristic approach is to scale the input images to the resolution that has also been used to collect the aesthetic ratings for the images (if applicable). Note that even a high-resolution image will automatically be downscaled to the maximum resolution of the display when displayed. This kind of scaling would lead to sufficiently small images (common display size of $1920 \times 1080$ pixels) for which the image properties can be calculated in a reasonable amount of time. Also note that resizing has a strong effect on many image properties, as described above. One and the same image may yield very different values for a given image property at different resolutions. The Github version also contains a script-only (no GUI) version of the Toolbox that can be used for local multi-threading or deployment to an HPC cluster.

In addition to calculating image properties, the toolbox provides a graphical interface for many common image pre-processing methods. These include:(a) image cropping, (b) image padding, (c) image color rotation, and (d) many image resizing options. For example, the methods can be used to resize all images to a similar image size by resizing the longer side of rectangular images to 512 or 1024 pixels and adjusting the shorter side accordingly to maintain the image ratio. Note that some image properties require a fixed type of image pre-processing that is always performed when calculating these properties (e.g., [P-]HOG-based Complexity or Edge-orientation entropy; see the online documentation for these image properties). For example, the image resolution is set to a fixed size of $512 \times 512$ pixels for the input image of CNN-based algorithms. Therefore, when calculating these image properties, these peculiarities should be taken into account in the choice of image pre-processing.

Furthermore, the individual image properties have very different computational demand. Image properties such as the simple mean values or standard deviations of the color channels are relatively fast to compute. The measures based on CNN features and the PHOG measures (Complexity, Anisotropy, and Self-similarity) are computationally more intensive.

By far the most demanding measure is the 2nd-order EOE, which compares the edge orientation of each edge in the image with all other edges (for the 10,000 strongest edges).

The revised Python 3 scripts of the Toolbox are partially optimized for runtime. For EOE, there is a separate additional C++ implementation in Cython. This version can easily achieve a speedup factor of $> 100$ on modern machines ( https://github.com/RBartho/C-version-2nd-Order-Edge-Orientation-Entropy).

## Discussion

The study of quantitative (objective) image properties and their role in aesthetic evaluations by human observers has been one of the central research topics in the field of experimental and computational aesthetics during the last 20 years or so. For the Aesthetics Toolbox, we selected 43 of these image properties and make algorithms to calculate them accessible to a broader audience. One of our aims is to enable researchers with no or little programming experience to calculate the image properties. To facilitate its usage, the entire Toolbox is written in one programming language (Python 3) and is built as a web application with the streamlit package (https://streamlit.io).

A major benefit of the newly created Toolbox is its simple-to-use interface, which runs platform-independent on most devices. The interface is intuitive and allows users to select images from the local hard drive for calculation, set the desired image properties and parameters, and export a single CSV file with the complete results. The toolbox can be used as a cloud version, but local installations are also possible. In particular, for local installation, the browser is used only as an interface and the application runs only on the local computer. No data is uploaded to the Internet and no external server is involved. This feature helps to avoid the uploading of large images, which could put a heavy load on the available bandwidth, and to get around copyrights issues that could possibly arise if protected images are uploaded to external servers.

The Aesthetics Toolbox includes image properties that cover widely different aspects of vision. Nonetheless, it still provides only a fraction of the image properties that have been studied in aesthetics research to date. Also, there are many different concepts for capturing aspects of vision that are rather divergent even if they carry the same name. For example, the methods for measuring symmetry in the Toolbox mirror rather different aspects of images and should be compared with caution. By the same token, Madan et al. (2018) found coefficients of correlations between different complexity measures ranging from 0.60 to 0.82. Such values are relatively low and raise the question of whether studies that use different complexity measures really analyze the same visual phenomenon (Madan et al., 2018; Van Geert & Wagemans, 2020, 2021). Moreover, calculated (objective) complexity does not always reflect the subjective impression of complexity of beholders (Forsythe et al., 2011; Marin & Leder, 2013; Nath et al., 2024), which can be influenced by subjective factors, such as familiarity with the stimuli or affect (Madan et al., 2018; Marin & Leder, 2013; McCormack & Gambardella, 2022).

The way in which the different measures are implemented adds another layer of complexity. During the development of the Toolbox, it became evident that simple implementation details can have a large impact on the calculated values of the individual image properties. Such details include image pre-processing (resizing and cropping), file formats (effect of different RGB color formats or non-lossless image compression), differences in the custom functions of the programming languages used, and even variations in the different versions of a single Python package used. From a scientific point of view, this situation is unfortunate because it makes it difficult to compare research results from different groups. To reach an agreement on common measures (and their consistent implementation) across the diverse research communities seems out of reach in the near future. Possible reasons for this situation are the lack of open access to used scripts, difficulties in using scripts, lack of programming knowledge, costly software licenses for proprietary code, or lack of maintenance of the code.

The Aesthetics Toolbox is our contribution to improve this situation. On the one hand, it provides the research community with a user-friendly tool to calculate a large selection of image properties. On the other hand, the Toolbox is designed as an open-source project. Other scientists are hereby invited to add their concepts and scripts to the Toolbox and to further develop it, be it by adding more image properties, or by integrating additional functionality. To facilitate such contributions, the source code for the entire Toolbox is available under MIT licenses on Github (https://github.com/RBartho/Aesthetics-Toolbox). The Github project page also contains detailed installation instructions for the most popular operating systems (Windows, Mac and Linux).

The Aesthetics Toolbox will eventually be accompanied by a searchable list of relevant datasets for aesthetics research (Koßmann, Bartho, Redies and Wagemans, unpublished data). This list will contain comprehensive information about each dataset and will allow researchers to find suitable datasets for their research projects using simple search criteria.

## Acknowledgments

The authors are grateful to colleagues who provided valuable feedback on their experience with the Aesthetics Toolbox, which greatly helped to improve its functions.

## Declarations

**Funding.** This work was supported by institute funds from the Institute of Anatomy, Jena University Hospital, and by the European Union (ERC, GRAPPA, 101053925, awarded to Johan Wagemans).

**Conflicts of Interest/Competing interests.** The authors have no competing interests to declare that are relevant to the content of this article.

**Ethics approval**. Not applicable because this work is purely descriptive and does not involve human or animal experiments.

**Consent to participate:** Not applicable

**Consent for publication:** Not applicable

**Availability of data and materials:** Data and materials are freely available at github.com/RBartho/Aesthetics-Toolbox

**Code availability.** The complete code for the toolbox is available at github.com/RBartho/Aesthetics-Toolbox.

## References

Aitken, P. P. (1974). Judgments of pleasingness and interestingness as functions of visual complexity. *Journal of Experimental Psychology*, *103*(2), 240-244. https://doi.org/10.1037/h0036787

Aks, D. J., & Sprott, J. C. (1996). Quantifying aesthetic preference for chaotic patterns. *Empirical Studies of the Arts*, *14*(1), 1-16. https://doi.org/10.2190/6V31-7M9R-T9L5-CDG9

Alvarez-Ramirez, J., Ibarra-Valdez, C., Rodriguez, E., & Dagdug, L. (2008). 1/f-noise structures in Pollock's drip paintings. *Physica A*, *387*, 281-295.

Amirshahi, S. A., Hayn-Leichsenring, G. U., Denzler, J., & Redies, C. (2015). JenAesthetics subjective dataset: Analyzing paintings by subjective scores. *Lecture Notes in Computer Science*, *8925*, 3-19. https://doi.org/10.1007/978-3-319-16178-5_1

Amirshahi, S. A., Koch, M., Denzler, J., & Redies, C. (2012). PHOG analysis of self-similarity in esthetic images. *Proceedings of SPIE (Human Vision and Electronic Imaging XVII)*, *8291*, 82911J. https://doi.org/10.1117/12.911973

Arnheim, R. (1954). *Art and Visual Perception: A Psychology of the Creative Eye*. Berkeley, CA: University of California Press.

Bar, M., & Neta, M. (2006). Humans prefer curved visual objects. *Psychological Science*, *17*(8), 645-648. https://doi.org/10.1111/j.1467-9280.2006.01759.x

Bekhtereva, V., & Muller, M. M. (2017). Bringing color to emotion: The influence of color on attentional bias to briefly presented emotional images. *Cognitive, Affective, & Behavioral Neuroscience*, *17*(5), 1028-1047. https://doi.org/10.3758/s13415-017-0530-z

Bell, A. J., & Sejnowski, T. J. (1997). The "independent components" of natural scenes are edge filters. *Vision Research*, *37*(23), 3327-3338. https://doi.org/10.1016/S0042-6989(97)00121-1

Berlyne, D. E. (1970). Novelty, complexity, and hedonic value. *Perception & Psychophysics*, *8*(5), 279-286. https://doi.org/10.3758/BF03212593

Bertamini, M., Palumbo, L., Gheorghes, T. N., & Galatsidas, M. (2016). Do observers like curvature or do they dislike angularity? *British Journal of Psychology*, *107*(1), 154-178. https://doi.org/10.1111/bjop.12132

Bertamini, M., Silvanto, J., Norcia, A. M., Makin, A. D. J., & Wagemans, J. (2018). The neural basis of visual symmetry and its role in mid-level and high-level visual processing. *Annals of the New York Academy of Sciences*, *1426*(1), 111-126. https://doi.org/10.1111/nyas.13667

Bies, A. J., Blanc-Goldhammer, D. R., Boydston, C. R., Taylor, R. P., & Sereno, M. E. (2016). Aesthetic responses to exact fractals driven by physical complexity. *Frontiers in Human Neuroscience*, *10*, 210. https://doi.org/10.3389/fnhum.2016.00210

Bies, A. J., Boydston, C. R., Taylor, R. P., & Sereno, M. E. (2016). Relationship between fractal dimension and spectral scaling decay rate in computer-generated fractals. *Symmetry-Basel*, *8*(7), 66. https://doi.org/10.3390/sym8070066

Bode, C., Helmy, M., & Bertamini, M. (2017). A cross-cultural comparison for preference for symmetry: comparing British and Egyptians non-experts. *Psihologija*, *50*, 383-402. https://doi.org/10.2298/PSI1703383B

Bosch, A., Zisserman, A., & Munoz, X. (2007). Representing shape with a spatial pyramid kernel. *Proceedings of the 6th ACM International Conference on Image and Video Retrieval*, 401-408. https://doi.org/10.1145/1282280.1282340

Bowers, J. S., Malhotra, G., Dujmovic, M., Montero, M. L., Tsvetkov, C., Biscione, V., Puebla, G., Adolfi, F., Hummel, J. E., Heaton, R. F., Evans, B. D., Mitchell, J., & Blything, R. (2022). Deep problems with neural network models of human vision. *Behavioral and Brain Sciences*, *46*, e385. https://doi.org/10.1017/S0140525X22002813

Brachmann, A., Barth, E., & Redies, C. (2017). Using CNN features to better understand what makes visual artworks special. *Frontiers in Psychology*, *8*, 830. https://doi.org/10.3389/fpsyg.2017.00830

Brachmann, A., & Redies, C. (2016). Using convolutional neural network filters to measure left-right mirror symmetry in images. *Symmetry*, *8*, 144. https://doi.org/10.3390/sym8120144

Brachmann, A., & Redies, C. (2017a). Computational and experimental approaches to visual aesthetics. *Frontiers in Computational Neuroscience*, *11*, 102. https://doi.org/10.3389/fncom.2017.00102

Brachmann, A., & Redies, C. (2017b). *Defining self-similarity of images using features learned by convolutional neural networks* Electronic Imaging, Human Vision and Electronic Imaging 2017, Burlingame, CA.

Braun, J., Amirshahi, S. A., Denzler, J., & Redies, C. (2013). Statistical image properties of print advertisements, visual artworks and images of architecture. *Frontiers in Psychology*, *4*, 808. https://doi.org/10.3389/fpsyg.2013.00808

Burton, G. J., & Moorhead, I. R. (1987). Color and spatial structure in natural scenes. *Applied Physics*, *26*, 157-170.

Canny, J. (1986). A computational approach to edge detection. *IEEE Transactions on Pattern Analysis and Machine Intelligence*, *PAMI-8*(6), 679-698.

Chamberlain, R. (2022). The interplay of objective and subjective factors in empirical aesthetics. In B. Ionescu, W. A. Bainbridge, & N. Murray (Eds.), *Human Perception of Visual Information* (pp. 115-132). Cham: Springer.

Chatterjee, A., & Vartanian, O. (2014). Neuroaesthetics. *Trends in Cognitive Sciences*, *18*(7), 370-375. https://doi.org/10.1016/j.tics.2014.03.003

Christensen, J. F., & Calvo-Merino, B. (2013). Dance as a subject for empirical aesthetics. *Psychology of Aesthetics, Creativity, and the Arts*, *7*(1), 76-88. https://doi.org/10.1037/a0031827

Chuquichambi, E. G., Vartanian, O., Skov, M., Corradi, G. B., Nadal, M., Silvia, P. J., & Munar, E. (2022). How universal is preference for visual curvature? A systematic review and meta-analysis. *Annals of the New York Academy of Sciences*, *1518*(1), 151-165. https://doi.org/10.1111/nyas.14919

Dalal, N., & Triggs, B. (2005). Histograms of oriented gradients for human detection. International Conference on Computer Vision & Pattern Recognition, 2, 886-893. https://doi.org/10.1109/CVPR.2005.177

Damiano, C., Wilder, J., Zhou, E. Y., Walther, D. B., & Wagemans, J. (2023). The role of local and global symmetry in pleasure, interest, and complexity judgments of natural scenes. *Psychology of Aesthetics, Creativity, and the Arts*, *17*(3), 322-337. https://doi.org/10.1037/aca0000398

Datta, R., Joshi, D., Li, J., & Wang, J. Z. (2006). Studying aesthetics in photographic images using a computational approach. *Lecture Notes in Computer Science*, *3953*, 288-301. https://doi.org/10.1007/11744078_23

Fairbanks, M. S., & Taylor, R. P. (2010). Measuring the scaling properties of temporal and spatial patterns: From the human eye to the foraging albatross. In *Nonlinear dynamical systems analysis for the behavioral sciences using real data*. Boca Raton: CRC Press.

Farzanfar, D., & Walther, D. B. (2023). Changing what you like: Modifying contour properties shifts aesthetic valuations of scenes. *Psychological Science*, *34*(10), 1101-1120. https://doi.org/10.1177/09567976231190546

Fernandez, D., & Wilkins, A. J. (2008). Uncomfortable images in art and nature. *Perception*, *37*(7), 1098-1113. https://doi.org/10.1068/p5814

Field, D. J. (1987). Relations between the statistics of natural images and the response properties of cortical cells. *Journal of the Optical Society of America A. Image Science and Vision*, *4*(12), 2379-2394.

Forsythe, A., Nadal, M., Sheehy, N., Cela-Conde, C. J., & Sawey, M. (2011). Predicting beauty: fractal dimension and visual complexity in art. *British Journal of Psychology*, *102*(1), 49-70. https://doi.org/10.1348/000712610X498958

Geisler, W. S., Perry, J. S., Super, B. J., & Gallogly, D. P. (2001). Edge co-occurrence in natural images predicts contour grouping performance. *Vision Research*, *41*(6), 711-724. https://doi.org/10.1016/S0042-6989(00)00277-7

Geller, H. A., Bartho, R., Thommes, K., & Redies, C. (2022). Statistical image properties predict aesthetic ratings in abstract paintings created by neural style transfer. *Frontiers in Neuroscience*, *16*, 999720. https://doi.org/10.3389/fnins.2022.999720

Gómez-Puerto, G., Munar, E., & Nadal, M. (2016). Preference for curvature: A historical and conceptual framework. *Frontiers in Human Neuroscience*, *9*, 712. https://doi.org/10.3389/fnhum.2015.00712

Graham, D. J., & Field, D. J. (2007). Statistical regularities of art images and natural scenes: spectra, sparseness and nonlinearities. *Spatial Vision*, *21*(1-2), 149-164. https://doi.org/10.1163/156856807782753877

Graham, D. J., & Field, D. J. (2008). Variations in intensity statistics for representational and abstract art, and for art from the Eastern and Western hemispheres. *Perception*, *37*(9), 1341-1352.

Graham, D. J., & Redies, C. (2010). Statistical regularities in art: Relations with visual coding and perception. *Vision Research*, *50*(16), 1503-1509. https://doi.org/10.1016/j.visres.2010.05.002

Grebenkina, M., Brachmann, A., Bertamini, M., Kaduhm, A., & Redies, C. (2018). Edge orientation entropy predicts preference for diverse types of man-made images. *Frontiers in Neuroscience*, *12*, 678. https://doi.org/10.3389/fnins.2018.00678

Güçlütürk, Y., Jacobs, R. H., & van Lier, R. (2016). Liking versus complexity: Decomposing the inverted U-curve. *Frontiers in Human Neuroscience*, *10*, 112. https://doi.org/10.3389/fnhum.2016.00112

Hekkert, P. (2006). Design aesthetics: principles of pleasure in design. *Psychology Science*, *48*, 157-172.

Hübner, R. (2022). Position biases in sequential location selection: Effects of region, choice history, and visibility of previous selections. *PLoS One*, *17*(10), e0276207. https://doi.org/10.1371/journal.pone.0276207

Hübner, R., & Fillinger, M. G. (2016). Comparison of objective measures for predicting perceptual balance and visual aesthetic preference. *Frontiers in Psychology*, *7*, 335. https://doi.org/10.3389/fpsyg.2016.00335

Hyvärinen, A., & Hoyer, P. O. (2001). A two-layer sparse coding model learns simple and complex cell receptive fields and topography from natural images. *Vision Research*, *41*, 2413-2423.

Ibarra, F. F., Kardan, O., Hunter, M. R., Kotabe, H. P., Meyer, F. A. C., & Berman, M. G. (2017). Image feature types and their predictions of aesthetic preference and naturalness. *Frontiers in Psychology, 8*, 632.

https://doi.org/10.3389/fpsyg.2017.00632

Iigaya, K., Yi, S., Wahle, I. A., Tanwisuth, K., & O'Doherty, J. P. (2021). Aesthetic

preference for art can be predicted from a mixture of low- and high-level visual

features. *Nature Human Behaviour, 5*(6), 743-755. https://doi.org/10.1038/s41562-

021-01124-6

Jacobsen, T., & Höfel, L. (2003). Descriptive and evaluative judgment processes: behavioral and electrophysiological indices of processing symmetry and aesthetics. *Cognitive,

Affective, & Behavioral Neuroscience, 3*(4), 289-299.

Joye, Y. (2007). Architectural lessons from environmental psychology: the case of biophilic architecture. *Review of General Psychiatry, 11*, 305-328.

Kersten, D. (1987). Predictability and redundancy of natural images. *Journal of the Optical

Society of America, Series A, 4*(12), 2395-2400.

http://www.ncbi.nlm.nih.gov/pubmed/3430226

Koch, M., Denzler, J., & Redies, C. (2010). 1/f2 Characteristics and isotropy in the Fourier

power spectra of visual art, cartoons, comics, mangas, and different categories of

photographs. *PLoS One, 5*(8), e12268. https://doi.org/10.1371/journal.pone.0012268

Kriegeskorte, N. (2015). Deep neural networks: A new framework for modeling biological vision and brain information processing. *Annual Review in Vision Science, 1*, 417-446.

https://doi.org/10.1146/annurev-vision-082114-035447

Krizhevsky, A., Sutskever, I., & Hinton, G. E. (2012). Imagenet classification with deep

convolutional neural networks. *Advances in Neural Information Processing Systems,

25*, 1097-1105.

Latto, R. (1995). The brain of the beholder. In R. L. Gregory, J. Harris, P. Heard, & D. Rose

(Eds.), *The artful eye* (pp. 66-94). New York, NY: Oxford University Press.

Lecun, B., & Bengio, Y. (1995). Convolutional networks for images, speech, and time series. *Handbook of Brain Theory and Neural Networks*, *3361*, 1995.

Leder, H., Belke, B., Oeberst, A., & Augustin, D. (2004). A model of aesthetic appreciation and aesthetic judgments. *British Journal of Psychology*, *95*(4), 489-508. https://doi.org/10.1348/0007126042369811

Leder, H., Tinio, P. P., Brieber, D., Kröner, T., Jacobsen, T., & Rosenberg, R. (2019). Symmetry is not a universal law of beauty. *Empirical Studies of the Arts*, *37*(1), 104-114. https://doi.org/10.1177/02762374187779

Li, C., & Chen, T. (2009). Aesthetic visual quality assessment of paintings. *IEEE Journal of Selected Topics in Signal Processing*, *3*(2), 236-252.

Li, J., Datta, R., Joshi, D., & Wang, J. (2006). Studying aesthetics in photographic images using a computational approach. *Lecture Notes in Computer Science*, *3953*, 288-301.

Li, R., & Zhang, J. (2020). Review of computational neuroaesthetics: bridging the gap between neuroaesthetics and computer science. *Brain Informatics*, *7*, 16. https://doi.org/10.1186/s40708-020-00118-w

Lindsay, G. W. (2020). Convolutional Neural Networks as a model of the visual system: Past, present, and future. *Journal of Cognitive Neuroscience*, *33*, 2017-2031. https://doi.org/10.1162/jocn_a_01544

Liu, J., Slota, G., Zheng, G., Wu, Z., Park, M., Lee, S., Rauschert, I., & Liu, Y. (2013). Symmetry Detection from Real World Images Competition 2013: Summary and Results. IEEE Computer Society Conference on Computer Vision and Pattern Recognition Workshops (CVPRW) 200-205.

Locher, P. J., Stappers, P. J., & Overbeeke, K. (1999). An empirical evaluation of the visual rightness theory of pictorial composition. *Acta Psychologica*, *103*, 261-280.

Luo, Y., & Tang, X. (2008). Photo and video quality evaluation: Focusing on the subject. *Computer Vision – ECCV 2008, Lecture Notes in Computer Science*, *5304*, 386-399.

Lyssenko, N., Redies, C., & Hayn-Leichsenring, G. U. (2016). Evaluating abstract art: Relation between term usage, subjective ratings, image properties and personality traits. *Frontiers in Psychology*, *7*, 973. https://doi.org/10.3389/fpsyg.2016.00973

Machado, P., Romero, J., Nadal, M., Santos, A., Correia, J., & Carballal, A. (2015). Computerized measures of visual complexity. *Acta Psychologica (Amsterdam)*, *160*, 43-57. https://doi.org/10.1016/j.actpsy.2015.06.005

Madan, C. R., Bayer, J., Gamer, M., Lonsdorf, T. B., & Sommer, T. (2018). Visual complexity and affect: Ratings reflect more than meets the eye. *Frontiers in Psychology*, *8*. https://doi.org/10.3389/fpsyg.2017.02368

Mallon, B., Redies, C., & Hayn-Leichsenring, G. U. (2014). Beauty in abstract paintings: Perceptual contrast and statistical properties. *Frontiers in Human Neuroscience*, *8*, 161. https://doi.org/10.3389/fnhum.2014.00161

Mandelbrot, B. (1983). *The fractal geometry of nature*. San Francisco: W. H. Freeman.

Marin, M. M., & Leder, H. (2013). Examining complexity across domains: Relating subjective and objective measures of affective environmental scenes, paintings and music. *PLoS One*, *8*(8), e72412. https://doi.org/10.1371/journal.pone.0072412

Mather, G. (2014). Artistic adjustment of image spectral slope. *Art & Perception*, *2*, 11-22.

Mather, G. (2018). Visual image statistics in the history of Western art. *Art & Perception*, *6*(2-3), 97-115. https://doi.org/10.1163/22134913-20181092

Mayer, S., & Landwehr, J. R. (2018). Quantifying visual aesthetics based on processing fluency theory: Four algorithmic measures for antecedents of aesthetic preferences. *Psychology of Aesthetics, Creativity, and the Arts*, *12*, 399-431. https://doi.org/10.1037/aca0000187

McCormack, J., & Gambardella, C. C. (2022). Complexity and aesthetics in generative and evolutionary art. *Genetic Programming and Evolvable Machines*, *23*(4), 535-556. https://doi.org/10.1007/s10710-022-09429-9

McManus, I. C., Stöver, K., & Kim, D. (2011). Arnheim's Gestalt theory of visual balance: Examining the compositional structure of art photographs and abstract images. *i-Perception*, *2*, 615-647.

Mehrotra, R., Namuduri, K. R., & Ranganathan, N. (1992). Gabor filter-based edge-detection. *Pattern Recognition*, *25*(12), 1479-1494. https://doi.org/10.1016/0031-3203(92)90121-X

Melmer, T., Amirshahi, S. A., Koch, M., Denzler, J., & Redies, C. (2013). From regular text to artistic writing and artworks: Fourier statistics of images with low and high aesthetic appeal. *Frontiers in Human Neuroscience*, *7*, 106. https://doi.org/10.3389/fnhum.2013.00106

Mureika, J. R., & Taylor, R. P. (2013). The abstract expressionists and Les Automatistes: A shared multi-fractal depth? *Signal Processing*, *93*, 573–578.

Murray, N., Marchesotti, L., & Perronnin, F. (2012). AVA: A large-scale database for aesthetic visual analysis. IEEE Conference on Computer Vision and Pattern Recognition (CVPR), 2408-2415.

Nakauchi, S., Kondo, T., Kinzuka, Y., Taniyama, Y., Tamura, H., Higashi, H., Hine, K., Minami, T., Linhares, J. M. M., & Nascimento, S. M. C. (2022). Universality and superiority in preference for chromatic composition of art paintings. *Scientific Reports*, *12*(1). https://doi.org/10.1038/s41598-022-08365-z

Nascimento, S. M., Linhares, J. M., Montagner, C., Joao, C. A., Amano, K., Alfaro, C., & Bailao, A. (2017). The colors of paintings and viewers' preferences. *Vision Research*, *130*, 76-84. https://doi.org/10.1016/j.visres.2016.11.006

Nath, S. S., Brändle, F., Schulz, E., Dayan, P., & Brielmann, A. (2024). Relating objective complexity, subjective complexity, and beauty in binary pixel patterns. *Psychology of Aesthetics, Creativity, and the Arts*, *(advance online publication)*. https://doi.org/10.1037/aca0000657

O'Hare, L., & Hibbard, P. B. (2011). Spatial frequency and visual discomfort. *Vision Research*, *51*(15), 1767-1777. https://doi.org/10.1016/j.visres.2011.06.002

Olshausen, B. A., & Field, D. J. (1996). Natural image statistics and efficient coding. *Network*, *7*(2), 333-339. https://doi.org/10.1088/0954-898X/7/2/014

Otsu, N. (1979). A threshold selection method from gray-level histograms. *IEEE Transactions on Systems, Man, and Cybernetics*, *SMC-9*, 62-66.

Palmer, S. E., Schloss, K. B., & Sammartino, J. (2013). Visual aesthetics and human preference. *Annual Review of Psychology*, *64*, 77-107. https://doi.org/10.1146/annurev-psych-120710-100504

Peli, E. (1990). Contrast in complex images. *Journal of the Optical Society of America Series A*, *7*(10), 2032-2040. https://doi.org/10.1364/josaa.7.002032

Peng, Y. (2022). Athec: A Python library for computational aesthetic analysis of visual media in social science research. *Computational Communication Research*, *4.1*, 323-349. https://doi.org/10.5117CCR2022.1.009.PENG

Rafegas, I., & Vanrell, M. (2016). Color spaces emerging from deep convolutional networks. Color and Imaging Conference, 2016(1), 225-230. https://doi.org/10.2352/ISSN.2169-2629.2017.32.225

Rafegas, I., & Vanrell, M. (2017). Color representation in CNNs: parallelisms with biological vision. Proceedings of the IEEE International Conference on Computer Vision Workshops, 2697-2705. https://doi.org/10.1109/ICCVW.2017.318

Redies, C. (2007). A universal model of esthetic perception based on the sensory coding of natural stimuli. *Spatial Vision*, *21*(1-2), 97-117. https://doi.org/10.1163/156856807782753886

Redies, C. (2015). Combining universal beauty and cultural context in a unifying model of visual aesthetic experience. *Frontiers in Human Neuroscience*, *9*, 219. https://doi.org/10.3389/fnhum.2015.00218

Redies, C., Amirshahi, S. A., Koch, M., & Denzler, J. (2012). PHOG-derived aesthetic measures applied to color photographs of artworks, natural scenes and objects. *ECCV 2012 Ws/Demos, Part I, Lecture Notes in Computer Science*, *7583*, 522–531. https://doi.org/10.1007/978-3-642-33863-2_54

Redies, C., & Brachmann, A. (2017). Statistical image properties in large subsets of traditional art, Bad Art, and abstract art. *Frontiers in Neuroscience*, *11*, 593. https://doi.org/10.3389/fnins.2017.00593

Redies, C., Brachmann, A., & Hayn-Leichsenring, G. U. (2015). Changes of statistical properties during the creation of graphic artworks. *Art & Perception*, *3*, 93-116. https://doi.org/10.1163/22134913-00002017

Redies, C., Brachmann, A., & Wagemans, J. (2017). High entropy of edge orientations characterizes visual artworks from diverse cultural backgrounds. *Vision Research*, *133*, 130-144. https://doi.org/10.1016/j.visres.2017.02.004

Redies, C., Grebenkina, M., Mohseni, M., Kaduhm, A., & Dobel, C. (2020). Global image properties predict ratings of affective pictures. *Frontiers in Psychology*, *11*, 953. https://doi.org/10.3389/fpsyg.2020.00953

Redies, C., & Gross, F. (2013). Frames as visual links between paintings and the museum environment: an analysis of statistical image properties. *Frontiers in Psychology*, *4*, 831. https://doi.org/10.3389/fpsyg.2013.00831

Redies, C., Hasenstein, J., & Denzler, J. (2007). Fractal-like image statistics in visual art: similarity to natural scenes. *Spatial Vision*, *21*(1-2), 137-148. https://doi.org/10.1163/156856807782753921

Ruderman, D. L. (1994). The statistics of natural scenes. *Network*, *5*, 517-548.

Schifanella, R., Redi, M., Aiello, L.M. (2015). An image is worth more than a thousand favorites: Surfacing the hidden beauty of flickr pictures. *Proceedings of the*

*International AAAI Conference on Web and Social Media*, *9*, 397-406.
https://doi.org/10.1609/icwsm.v9i1.14612

Shannon, C. E. (1948). A mathematical theory of communication. *Bell System Technical Journal*, *27*(4), 623-656. https://doi.org/10.1002/j.1538-7305.1948.tb00917.x

Sidhu, D. M., McDougall, K. H., Jalava, S. T., & Bodner, G. E. (2018). Prediction of beauty and liking ratings for abstract and representational paintings using subjective and objective measures. *PLoS One*, *13*(7), e0200431.
https://doi.org/10.1371/journal.pone.0200431

Simoncelli, E. P., & Olshausen, B. A. (2001). Natural image statistics and neural representation. *Annual Review of Neuroscience*, *24*, 1193-1216.

Spehar, B., Clifford, C. W. G., Newell, B. R., & Taylor, R. P. (2003). Universal aesthetic of fractals. *Computers & Graphics*, *27*, 813-820. https://doi.org/10.1016/S0097-8493(03)00154-7

Spehar, B., & Taylor, R. P. (2013). Fractals in art and nature: Why do we like them? S&T/SPIE Electronic Imaging, Burlingame, California, United States, 8651, 865118.
https://doi.org/10.1117/12.2012076

Spehar, B., Walker, N., & Taylor, R. P. (2016). Taxonomy of individual variations in aesthetic responses to fractal patterns. *Frontiers in Human Neuroscience*, *10*, 350.
https://doi.org/10.3389/fnhum.2016.00350

Spehar, B., Wong, S., van de Klundert, S., Lui, J., Clifford, C. W. G., & Taylor, R. P. (2015). Beauty and the beholder: the role of visual sensitivity in visual preference. *Frontiers in Human Neuroscience*, *9*. https://doi.org/10.3389/fnhum.2015.00514

Stanischewski, S., Altmann, C. S., Brachmann, A., & Redies, C. (2020). Aesthetic perception of line patterns: effect of edge-orientation entropy and curvilinear shape. *i-Perception*, *11*(5), 2041669520950749.

Taylor, R. P. (2002). Order in Pollock's chaos - Computer analysis is helping to explain the appeal of Jackson Pollock's paintings. *Scientific American*, *287*(6), 116-121. https://doi.org/10.1038/scientificamerican1202-116

Taylor, R. P., Micolich, A. P., & Jonas, D. (1999). Fractal analysis of Pollock's drip paintings. *Nature*, *399*, 422.

Taylor, R. P., Newell, B., Spehar, B., & Clifford, C. W. G. (2005). Fractals: a resonance between art and nature. In M. Emmer (Ed.), *The Visual Mind II* (pp. 53-63). Cambridge: MIT Press.

Taylor, R. P., Spehar, B., Van Donkelaar, P., & Hagerhall, C. M. (2011). Perceptual and physiological responses to Jackson Pollock's fractals. *Frontiers in Human Neuroscience*, *5*, 60. https://doi.org/10.3389/fnhum.2011.00060

Taylor, R. P., Spehar, B., Wise, J. A., Clifford, C. W., Newell, B. R., Hagerhall, C. M., Purcell, T., & Martin, T. P. (2005). Perceptual and physiological responses to the visual complexity of fractal patterns. *Nonlinear Dynamics, Psychology, and Life Sciences*, *9*(1), 89-114.

Taylor, R. P., & Sprott, J. C. (2008). Biophilic fractals and the visual journey of organic screen-savers. *Nonlinear Dynamics, Psychology, and Life Sciences*, *12*, 117-129.

Thieleking, R., Medawar, E., Disch, L., & Witte, A. V. (2020). Art. pics database: An open access database for art stimuli for experimental research. *Frontiers in Psychology*, *11*, 3537.

Thömmes, K., & Hübner, R. (2018). Instagram likes for architectural photos can be predicted by quantitative balance measures and curvature [Original Research]. *Frontiers in Psychology*, *9*(1050). https://doi.org/10.3389/fpsyg.2018.01050

Tinio, P. P. L., Leder, H., & Strasser, M. (2011). Image quality and the aesthetic judgment of photographs: Contrast, sharpness, and grain teased apart and put together. *Psychology* *of Aesthetics, Creativity, and the Arts*, *5*(2), 165-176.

https://doi.org/10.1037/a0019542

Tolhurst, D. J., Tadmor, Y., & Chao, T. (1992). Amplitude spectra of natural images. *Ophthalmologic and Physiological Optics*, *12*(2), 229-232.

Tong, H., Li, M., Zhang, H.-J., He, J., & Zhang, C. (2004). Classification of digital photos taken by photographers or home users. *Lecture Notes in Computer Science*, *3331*, 198-205.

van Dongen, N. N. N., & Zijlmans, J. (2017). The science of art: The universality of the law of contrast. *American Journal of Psychology*, *130*(3), 283-294. <Go to ISI>://WOS:000407872700003

Van Geert, E., & Wagemans, J. (2020). Order, complexity, and aesthetic appreciation. *Psychology of Aesthetics, Creativity, and the Arts*, *14*(2), 135. https://doi.org/10.1037/aca0000224

Van Geert, E., & Wagemans, J. (2021). Order, complexity, and aesthetic preferences for neatly organized compositions. *Psychology of Aesthetics, Creativity, and the Arts*, *15*(3), 484-504. https://doi.org/10.1037/aca0000276

Viengkham, C., Isherwood, Z., & Spehar, B. (2019). Fractal-scaling properties as aesthetic primitives in vision and touch. *Axiomathes*, *32*, 869-888. https://doi.org/10.1007/s10516-019-09444-z

Viengkham, C., & Spehar, B. (2018). Preference for fractal-scaling properties across synthetic noise images and artworks. *Frontiers in Psychology*, *9*, 1439. https://doi.org/10.3389/fpsyg.2018.01439

Vissers, N., Moors, P., Genin, D., & Wagemans, J. (2020). Exploring the role of complexity, content and individual differences in aesthetic reactions to semi-abstract photographs. *Art & Perception*, *8*(1), 89-119. https://doi.org/10.1163/22134913-20191139

Vissers, N., & Wagemans, J. (2023). The photographer's visual grammar: Visual rightness and aesthetics of artistic photographs. *Art & Perception advance online publication*. https://doi.org/10.1163/22134913-bja10047

Vukadinovic, M., & Markovic, S. (2012). Aesthetic experience of dance performances. *Psihologija*, *45*(1), 23-41. https://doi.org/10.2298/Psi1201023v

Wagemans, J. (1995). Detection of visual symmetries. *Spatial Vision*, *9*(1), 9-32. https://doi.org/10.1163/156856895x00098

Wagemans, J. (1997). Characteristics and models of human symmetry detection. *Trends in Cognitive Sciences*, *1*(9), 346-352. https://doi.org/10.1016/S1364-6613(97)01105-4

Walther, D. B., Farzanfar, D., Han, S. H., & Rezanejad, M. (2023). The mid-level vision toolbox for computing structural properties of real-world images. *Frontiers in Computer Science*, *5*, 1140723. https://doi.org/10.3389/fcomp.2023.1140723

Watier, N. (2018). The saliency of angular shapes in threatening and nonthreatening faces. *Perception*, *47*(3), 306-329. https://doi.org/10.1177/0301006617750980

Watier, N. (2024). Measures of angularity in digital images. *Behavior Research Methods*, *(published online)*. https://doi.org/10.3758/s13428-024-02412-5

Wilson, A., & Chatterjee, A. (2005). The assessment of preference for balance: Introducing a new test. *Empirical Studies of the Arts*, *23*, 165-180.

Wundt, W. (1874). *Grundzüge der physiologischen Psychologie*. Leipzig: Wilhelm Engelmann Verlag.

Zhang, J., Miao, Y., & Yu, J. (2021). A comprehensive survey on computational aesthetic evaluation of visual art Images: Metrics and challenges. *IEEE Access*, *9*, 77164 - 77187. https://doi.org/10.1109/ACCESS.2021.3083075